\documentclass[journal]{IEEEtran}
\usepackage{graphicx}
\usepackage{epstopdf}
\usepackage{grffile}
\usepackage{amsthm}
\usepackage{array}
\usepackage[symbol]{footmisc}
\usepackage{algorithmicx}
\usepackage[ruled,vlined]{algorithm2e}
\usepackage{algpseudocode}
\usepackage{amssymb}
\usepackage{amsfonts}
\usepackage{mathrsfs}
\usepackage{amsmath}
\usepackage{txfonts}
\usepackage{autobreak} 
\usepackage{subfigure} 
\usepackage{cite} 
\usepackage{soul}
\usepackage{xcolor}
\usepackage{color}
\usepackage{mathtools}
\usepackage{bm}
\usepackage{enumitem}

\usepackage{multirow}
\usepackage{textcomp}
\usepackage{float}
\usepackage{cleveref}
\usepackage{booktabs}
\usepackage{colortbl}
\usepackage{stfloats}    

\algnewcommand\algorithmicforeach{\textbf{for each}}
\algdef{S}[FOR]{ForEach}[1]{\algorithmicforeach\ #1\ \algorithmicdo}

\begin{document}

\title{8-DoFs Cable Driven Parallel Robots for Bimanual Teleportation}
\author{Hung~Hon~Cheng$^1$ and Josie Hughes$^2$ ~\IEEEmembership{ Member,~IEEE}
\thanks{ 
H. H. Cheng (corrsponding author.) (hung.cheng@epfl.ch) and J. Hughes
 (josie.hughes@epfl.ch) are with the CREATE Lab, EPFL, Lausanne, Switzerland
       }%
}

\markboth{Journal of \LaTeX\ Class Files}%
{Shell \MakeLowercase{\textit{et al.}}: Bare Demo of IEEEtran.cls for IEEE Journals}

\maketitle
\begin{abstract}
Teleoperation plays a critical role in intuitive robot control and imitation learning, particularly for complex tasks involving mobile manipulators with redundant degrees of freedom (DoFs). However, most existing master controllers are limited to 6-DoF spatial control and basic gripper control, making them insufficient for controlling high-DoF robots and restricting the operator to a small workspace.
In this work, we present a novel, low-cost, high-DoF master controller based on Cable-Driven Parallel Robots (CDPRs), designed to overcome these limitations. The system decouples translation and orientation control, following a scalable 3 + 3 + n DoF structure: 3 DoFs for large-range translation using a CDPR, 3 DoFs for orientation using a gimbal mechanism, and n additional DoFs for gripper and redundant joint control. Its lightweight cable-driven design enables a large and adaptable workspace while minimizing actuator load. The end-effector remains stable without requiring continuous high-torque input, unlike most serial robot arms.
We developed the first dual-arm CDPR-based master controller using cost-effective actuators and a simple mechanical structure. In demonstrations, the system successfully controlled an 8-DoF robotic arm with a 2-DoF pan-tilt camera, performing tasks such as pick-and-place, knot tying, object sorting, and tape application. The results show precise, versatile, and practical high-DoF teleoperation.

\end{abstract}

\begin{IEEEkeywords}
Cable-driven parallel robot, teleoperation, robot manipulation, redundancy control
\end{IEEEkeywords}

\section{Introduction} \label{sec:intro}

Cable-Driven Parallel Robots (CDPRs) have primarily been used as large-scale and heavy-loading carrier due to their wide workspace, high payload capacity, and rigid end-effectors \cite{qian2018review}. However, their use as haptic devices \cite{poitrimol2025cable,fan2022development,xue2023new,poitrimol2020haptic} or teleoperation controllers \cite{park2021portable,kim2022remotely,yang2015haptic} has been limited since the concept was introduced in \cite{kawamura1993new}. This paper presents a novel design of 8-DoF dual CDPRs teleoperation system, using low-cost actuators, for precise bimanual manipulation. The proposed mechanical design resolves the limited orientation workspace problem of CDPRs and leverages the excellent force transmission, low inertia, and simple components of CDPRs for complex manipulation tasks.

As teleoperation can be widely used in robotic control applications, such as surgical systems and imitation learning, the main types of master controllers include serial robots (e.g. a kinematic linkage) \cite{silva2009phantom,fu2024mobile}, parallel platforms \cite{son2013human,dasgupta2000stewart,risiglione2021passivity}, and vision-based gesture recognizers \cite{bimbo2017teleoperation,cheng2024open}. As shown in \cite{samur2012performance}, the commercial master controller ranges from narrow workspace with 3-DoFs to large workspace up to 7-DoFs for different types of task motion. Physical controllers are often preferred for their haptic feedback, while vision-based controllers require additional wearable haptic devices \cite{meli2018hbracelet,li2019intuitive}, but they still can not provide intuitive spatial force feedback in translation and orientation. 

Serial linkage designs suffer from accumulated link mass and DoFs, which demands either high-performance actuators or lightweight materials, especially in applications requiring human-arm range workspace and heavy end-effector. The high torque actuator may require real-time control algorithms to compensate for mass inertia and ensure smooth operation \cite{hashtrudi2001analysis}.
In contrast, CDPRs shares the end-effector's weight and results in low inertia, a large and scalable workspace, and sufficient force feedback with simpler mechanism and control. These features make CDPRs especially well-suited for designing master controllers with extended reach and reduced mechanical complexity.

\begin{figure}[t]
    \centering
    \includegraphics[width = 1\columnwidth]{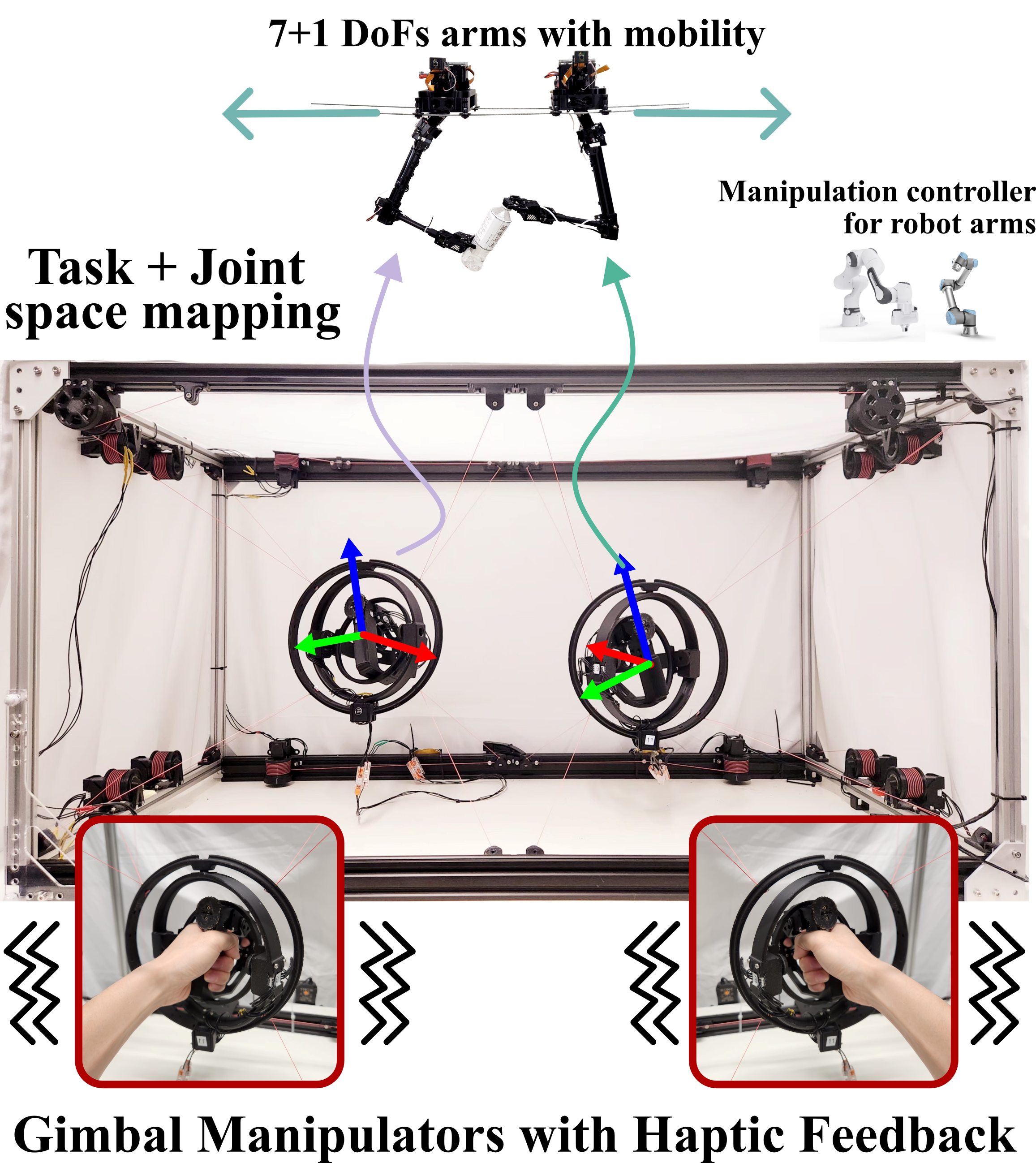}
    \caption{Two 8-DoF cable-driven parallel robots (CDPRs), each equipped with a gimbal-based controller, are used to provide decoupled translational and rotational teleoperation input for a mobile robotic arm.}
    \label{teleop_first_fig}
\end{figure}
Compared to serial-type controllers, few high-DoF CDPRs have been developed that support both translation and orientation control for bimanual applications. Most existing CDPRs are designed primarily for haptic feedback \cite{poitrimol2025cable,fan2022development,xue2023new,poitrimol2020haptic}, while others provide mainly translational motion with limited orientation capabilities \cite{kraus2015pulley,park2021portable,kim2022remotely,yang2015haptic}. This limitation stems from the fact that traditional CDPRs typically have a narrow, position-dependent orientation workspace. Although various end-effector designs \cite{metillon2021cable,zhang2019efficient,sun2022design} and reconfigurable mechanisms \cite{rodriguez2018cable,jamshidifar2020static,cheng2023cable} have been proposed to expand orientation capabilities, the translation and orientation workspaces remain coupled in most current systems. This coupling introduces the need for additional actuators and increases system complexity—issues directly addressed by the proposed decoupled design.

Additionally, most commercial teleoperation controllers offer a maximum of 7 DoFs, including the gripper, as they are typically designed for 7-DoF arms \cite{li2022novel,rebelo2014bilateral,guthart2000intuitive}. However, recent developments in high-DoF bimanual systems with mobility demand more control inputs, yet lack a suitable intuitive master interface \cite{zhou2024learning,liu2024hierarchical,dadiotis2023whole}. For example, in one application involving cable-traversing bimanual robotic arms for agricultural tasks \cite{cheng2025cafes}, a total of 8 DoFs per arm is required.
To address this gap, the proposed system introduces a $6+n$ DoF master controller, where the base 6 DoFs represent spatial translation and orientation, and the additional 
$n$ DoFs correspond to gripper and redundant joint or mobility control. For example, in the agricultural setup, these additional DoFs are used to control cable-based mobility and grasping of each individual arm.

This paper presents a new mechanical design for a teleoperation system based on Cable-Driven Parallel Robots (CDPRs), aiming to provide an intuitive, cost-effective, and scalable master controller. The system has 8 lightweight motors for 3-DoF translational motion via CDPR, 3 motors for gimbal-based orientation, and 2 motors for auxiliary control such as clutching. A simple foot pedal interface is also included for logic-level commands. 
These features reduce the mechanical complexity and actuator demands, enabling lightweight operation with low-cost components such as current-controlled Dynamixel motors.
Additionally, the system’s expandability allows users to increase the workspace simply by adjusting the anchoring points, without the need to redesign core mechanical components. This flexibility makes it suitable for applications requiring large-scale or high-DoF teleoperation, which are currently underserved by commercial solutions. Moreover, the decoupled design and $6+n$ DoF framework make the system particularly well-suited for dexterous and redundant manipulation tasks in both industrial and field robotics.

\section{Methods}
The proposed system consists of two main components for each hand as shown in \Cref{teleop_first_fig}: a 6-DoF CDPR actuated by 8 motors, and a 5-DoF gimbal manipulator mounted as its end-effector. While the CDPR can theoretically support 6 DoFs, its orientation is passively constrained by the manipulator design and cable anchor configuration. Therefore, during teleoperation, only translational motion is used from the CDPR, and orientation is fully handled by the gimbal. The following sections describe the CDPR and gimbal manipulator in detail.
In this setup, cable weight is considered negligible, and the cables are assumed to have low elasticity, providing high stiffness and minimal elongation during operation.

\subsection{Kinematics Model of 6-DoFs CDPR}
\begin{figure}[h]
    \centering
    \includegraphics[width = 0.23\textwidth]{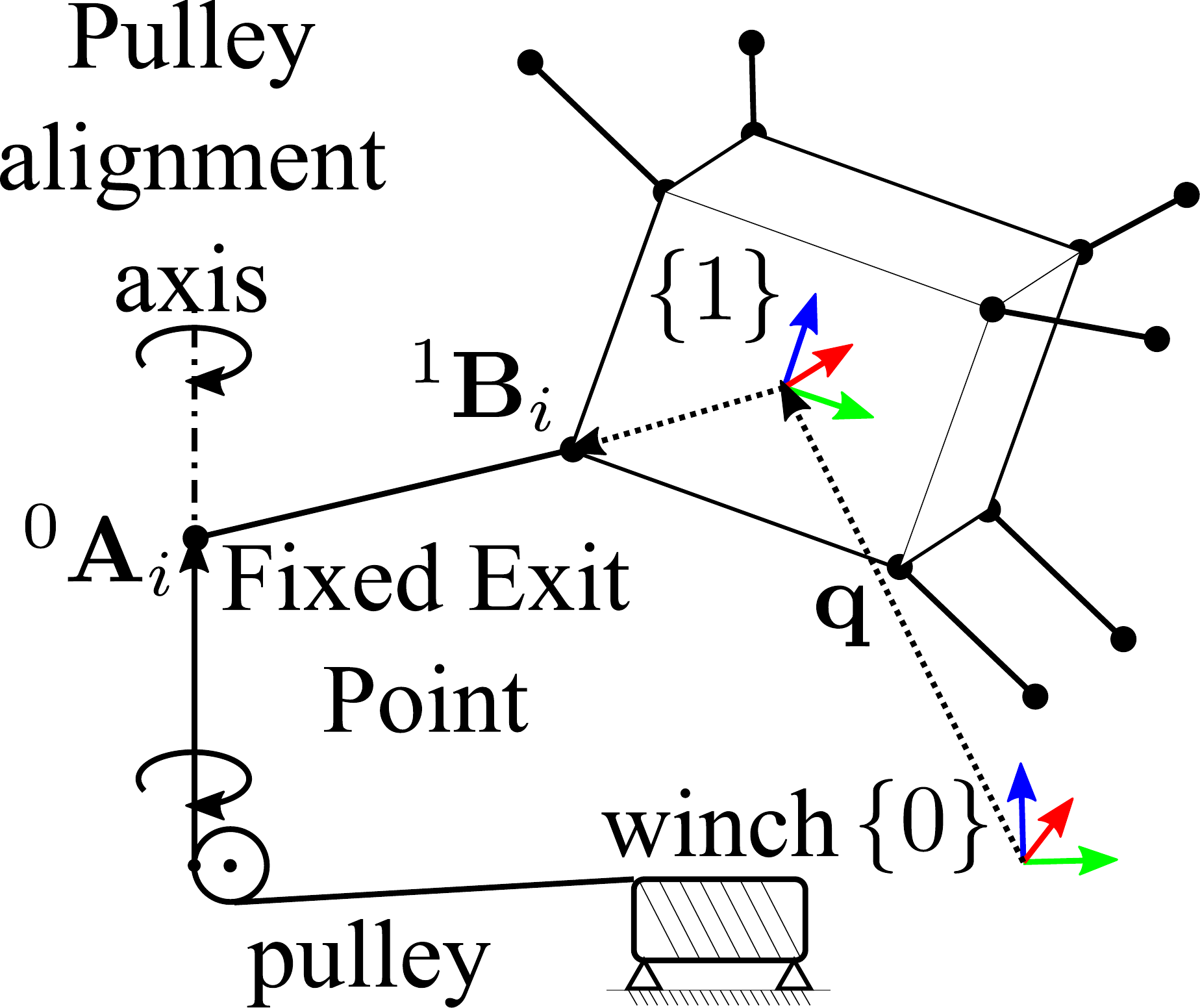}
    \caption{Kinematic of CDPRs}
    \label{fig_MODEL}
\end{figure}
The coordinates of 6-DoF, 8-cables CDPR's are expressed as $\mathbf{q} = [q_1, q_2, \dots, q_6]^T := [\mathbf{q}_t^T, \mathbf{q}_o^T]^T \in \mathbb{R}^6$, where $\mathbf{q}_t$ and $\mathbf{q}_o$ represent the translation and euler orientation of the end-effector, respectively. As shown in \Cref{fig_MODEL}, the kinematics of the CDPR is constructed by the cable segment $\mathbf{AB}_i$. It extends from the base cable attachment location ${}^0\mathbf{A}_{i} = [A_{ix}~A_{iy}~A_{iz}]^T$ (relative to the global frame $\{0\}$) to the end-effector cable attachment location ${}^1\mathbf{B}_{i} = [B_{ix}~B_{iy}~B_{iz}]^T$ (relative to local frame $\{1\}$) and can be expressed as:
\begin{align}
\label{equ:cableVector}
    \mathbf{AB}_i(\mathbf{q}) &=
    \mathbf{q}_t + {}^0_1R(\mathbf{q}_o) {}^1\mathbf{B}_i - {}^0\mathbf{A}_i \nonumber \\
    &=
    \begin{bmatrix}
        x - A_{ix} \\
        y - A_{iy} \\
        z - A_{iz}
    \end{bmatrix} +
    {}^0_1R(\mathbf{q}_o)
    \begin{bmatrix}
        B_{ix} \\
        B_{iy} \\
        B_{iz}
    \end{bmatrix}
\end{align}
with $\mathbf{q}_t = [x, y, z]^T$ indicates the $x$-$y$-$z$ coordinates, and $^0_1R$ is the rotation matrix between frames $\{0\}$ and $\{1\}$. In the proposed system, $^0_1R$ is the rotational matrix using Euler angles with the sequence $XYZ$, such as
\begin{equation}\label{equ_r_rotation}
^0_1R = R_x(q_4)R_y(q_5)R_z(q_6)
\end{equation}

In general, ${}^0\mathbf{A}{i}$ can be a function of $\mathbf{q}$ when pulley kinematics are considered. However, in the proposed design, ${}^0\mathbf{A}{i}$ are intentionally implemented as fixed point outlets. This simplification reduces mechanical complexity and allows the cable attachments to be modeled as static points, streamlining both the hardware and kinematic computation.

\subsection{Gimbal Manipulator with additional 2-DoFs}
\begin{figure}[h]
    \centering
    \includegraphics[width = 0.9\columnwidth]{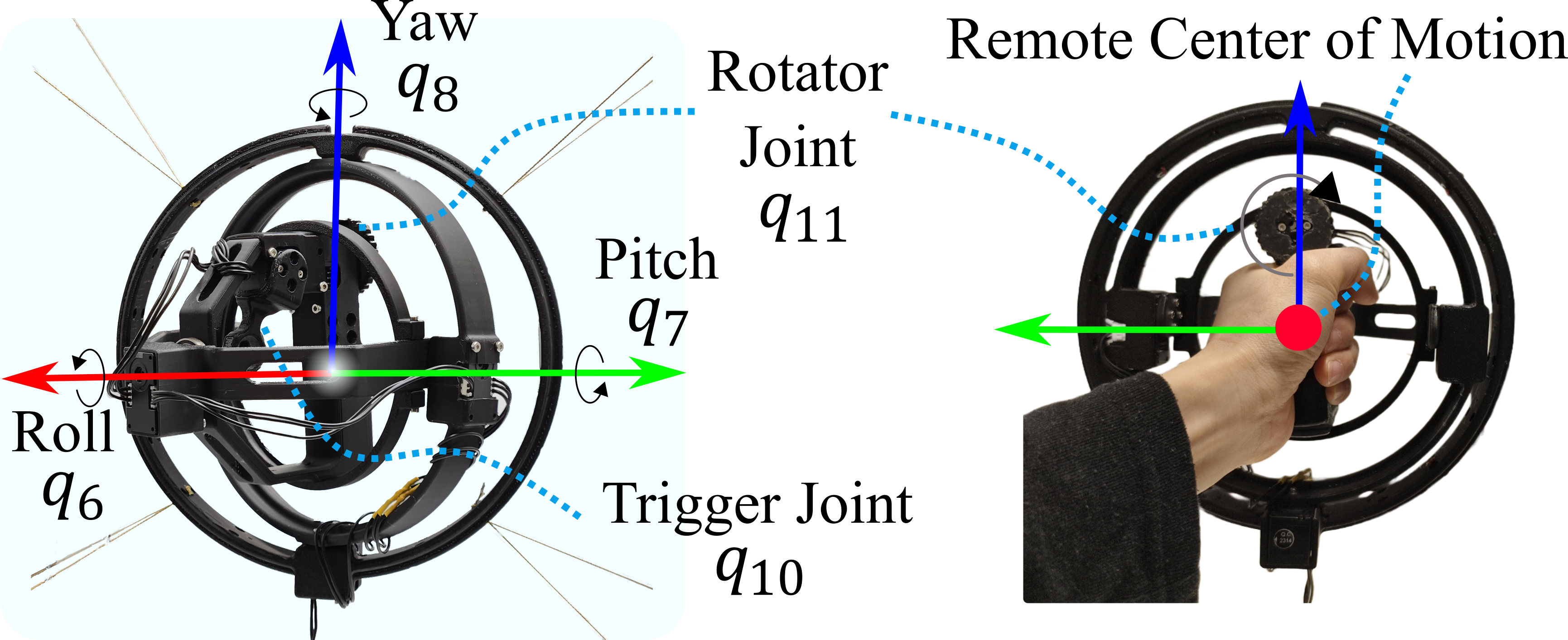}
    \caption{Gimbal Manipulator and its control variables. }
    \label{fig_gimbal}
\end{figure}
Due to the limited orientation range and wrench capability of the CDPR, a gimbal manipulator (shown in Fig. \ref{fig_gimbal}) is mounted at the end-effector to provide three-axis rotation in roll, pitch, and yaw. The Remote Center of Motion (RCM) design ensures that all three rotational axes intersect at a single control point, which also corresponds to the translational position $\mathbf{q}_t$ of the CDPR. In addition, two extra actuators are integrated into the gimbal to offer additional control inputs. The control output from the manipulator is represented as:
\begin{equation}
    \mathbf{q}_m = [q_7,q_8,q_9,q_{10},q_{11}]^T \in \mathbb{R}^5
\end{equation}
 where $q_7$, $q_8$, and $q_9$ represent the roll, pitch, and yaw angles, respectively, and $q_{10}$ and $q_{11}$ represent the trigger and a rotation joints.
The use of the trigger and rotation joints is user-defined. Generally, for teleoperation, the trigger joint can be used as the gripper angle controller, while the rotation joint can be used to adjust the teleoperation ratio, control redundant joints, or manage arbitrary DoFs.

More DoFs can be added to the gimbal manipulator which can be tailored by the specific application. Because the additional actuator's weight is shared by 8 cables, the effect of adding addition payload to the gimble is minimal on each actuator, unlike the serial robotic arms.

\subsection{Forward Kinematics}
Due to the low torque control and the backdrivability of the gimbal joints, the end-effector maintains a zero orientation in most of the workspace, as the operators can only apply linear force to the cables and cannot directly introduce torque. 
The Forward Kinematic(FK) problem can be solved using the Levenberg-Marquardt (LM) optimization method \cite{schmidt2013implementing, santos2021real}. It is formulated as 
\begin{equation}\label{equ_lm_q}
\begin{aligned}
    \mathbf{q}_t^* = \min_{\mathbf{q}} \quad & \| \mathbf{l} - \mathbf{l}_q(\mathbf{q}) \| \\
    \text{subject to} \quad & \mathbf{q}_{\min} \leq \mathbf{q} \leq \mathbf{q}_{\max}
\end{aligned}
\end{equation}
where $\mathbf{l} = [l_1, l_2, \cdots , l_8]^T$ represents the actual cable lengths and $\mathbf{l}_q(\mathbf{q})=[||\mathbf{AB}_1(\mathbf{q})||,||\mathbf{AB}_2(\mathbf{q})||,\cdots], \forall i$ is from \Cref{equ:cableVector}. However, it is well known that the measurement accuracy of the cable attachment point ${}^0\mathbf{A}_i$ and initial guess significantly affects the convergence and sensitivity of the LM method \cite{transtrum2012improvements}.
In practice, the actual cable length is obtained by adding the length change $\Delta l_i$ to the initial cable length $l_{i,0}$, such that $l_i = l_0 + \Delta l_i$ \cite{lau2018initial}. Therefore, accurate estimation of the initial cable length is essential, as discussed in \cite{lau2018initial}.

It is important to note that, due to the RCM design, the translational position of the end-effector remains accurate regardless of changes in its orientation. This inherent decoupling ensures that orientation variations do not affect the accuracy of the translation variable.
Additionally, as shown in the results section, the end-effector's orientation $\mathbf{q}_o$ varies only slightly, even when the initial pose guess is not highly accurate. In contrast, the translational displacement $\mathbf{q}_t = [x, y, z]^T$ remains consistently accurate. As a result, only the translational component is utilized from the forward kinematics solver.

\subsection{Overall Task Space Output}
Eventually, the master's 8-DoFs task space command can be obtained by
\begin{equation}\label{master_pose}
    \mathbf{X}_m =
        [\mathbf{q}_t^T - \mathbf{q}_{t,0}^T, \mathbf{q}_m^T ]^T \in \mathbb{R}^8
\end{equation}
where $\mathbf{q}_{t,0} = [x_0, y_0, z_0]^T$ is the reference position at the moment the user enables the task-space command $\mathbf{X}_m$. In the proposed system, $\mathbf{q}_{t,0}$ is recommended to be set at the center of the workspace, where the orientation $||\mathbf{q}_o|| = 0$.

It is important to highlight the role of the reference position. The workspace of the master and the teleoperated robot can differ, for example when the teleoperated robot is a serial manipulator or when its workspace is significantly larger than that of the master. In such cases, the position command $\mathbf{q}_t^T$ is not directly transferred to the teleoperated robot. Instead, only the angular components $\mathbf{q}_m^T$ are forwarded. The user can define the reference translation position as needed, but the orientation and other angle-related joints should follow the motion of the gimbal manipulator. This consideration is one of the main motivations behind the proposed design, which decouples the orientation and translation components.
For example, in the teleoperated robot side, the task space command will be 
\begin{equation}\label{slave_pose}
    \mathbf{X}_s =
        [\mathbf{x}_{o}+\mathbf{q}_t^T - \mathbf{q}_{t,0}^T, \mathbf{q}_m^T ]^T \in \mathbb{R}^8
\end{equation}
where $\mathbf{x}_o = [x_0, y_0, z_0]^T\neq\mathbf{q}_{t,0} $ is the reference position of the teleoperated robot before the motion is enabled.

Although the orientation of the end-effector $\mathbf{q}_o$ is not critical in most areas of the workspace due to the backdrivable gimbal manipulator, which prevents torque from being directly applied to the end-effector, there are still undesired regions where $\mathbf{q}_o$ becomes relatively large. In such cases, the operator may feel discomfort during teleoperation when $\mathbf{q}_o$ increases.

To address this issue, a virtual wall is implemented to define the near-zero-orientation region of the workspace and to alert the operator when movement may lead to a significant increase in $\mathbf{q}_o$. Haptic feedback is generated to remind the operator repositioning the reference point. In the proposed system, this near-zero-orientation workspace is identified using measurements from the OptiTrack system, ensuring that $||\mathbf{q}_o|| \leq 10^\circ$.

\subsection{Haptic Feedback}\label{sec:hapticfeedback}

Haptic feedback can be obtained from either virtual fixture or sensor feedback. The virtual fixture which is usually defined by both master and teleoperated robot's kinematic. It is usually defined as surfaces function such as cone, ball, and planes. 
Our virtual fixture is defined as the set of Cartesian positions $\mathbf{q}_t = [x, y, z]^T \in \mathbb{R}^3$ for which the corresponding end-effector orientation remains within a small angular threshold. Specifically, the robot's workspace is defined as
\begin{equation}
    \mathcal{W} = \left\{ \mathbf{q}_t \in \mathbb{R}^3 \ \middle| \ \|\mathbf{q}_t\| \leq 10^\circ \right\}
\end{equation}
and the virtual wall is defined as a ellipsoid 
\begin{equation}
    \mathcal{V}=\frac{(x - x_0)^2}{r_x^2} + \frac{(y - y_0)^2}{r_y^2} + \frac{(z - z_0)^2}{r_z^2} \leq 1
\end{equation}
Due to the fact that, when the end-effector moves outside the ellipsoidal virtual wall, the repulsive force (haptic feedback) can be simplified as a scalar gain $\sigma$ multiplied by a unit vector pointing toward the center of the workspace. This force can be expressed as
\begin{equation}
    \dot{\mathbf{q}}=\sigma \hat{\mathbf{q}} \in \mathbb{R}^3,\quad \textit{where}\quad\hat{\mathbf{q}}=\frac{\mathbf{q}_t - \mathbf{q}_{t,0}}{||\mathbf{q}_t - \mathbf{q}_{t,0}||}
\end{equation}
Finally, by solving the equation 
\begin{equation}
   \mathbf{V}_q= -\mathbf{J}(\mathbf{q})^T \mathbf{f}  
\end{equation}
where $\mathbf{V}_q = [\dot{\mathbf{q}}^T, 0, 0, 0]^T \in \mathbb{R}^6$, the cable force $\mathbf{f}$ (corresponding to motor current) can be determined. Since the purpose of the haptic feedback is to prompt the operator to relocate the reference point $\mathbf{q}_{t,0}$ rather than strictly constrain the robot’s movement outside the virtual fixture, a high-frequency vibration is applied to enhance the user's perception of the feedback. However, in our setup, the cable force and current mapping is only estimated by  the actuator's torque and the measured current, no additionally tension sensor is used. 
\begin{figure*}[t]
    \centering
    \includegraphics[width=0.97\textwidth]{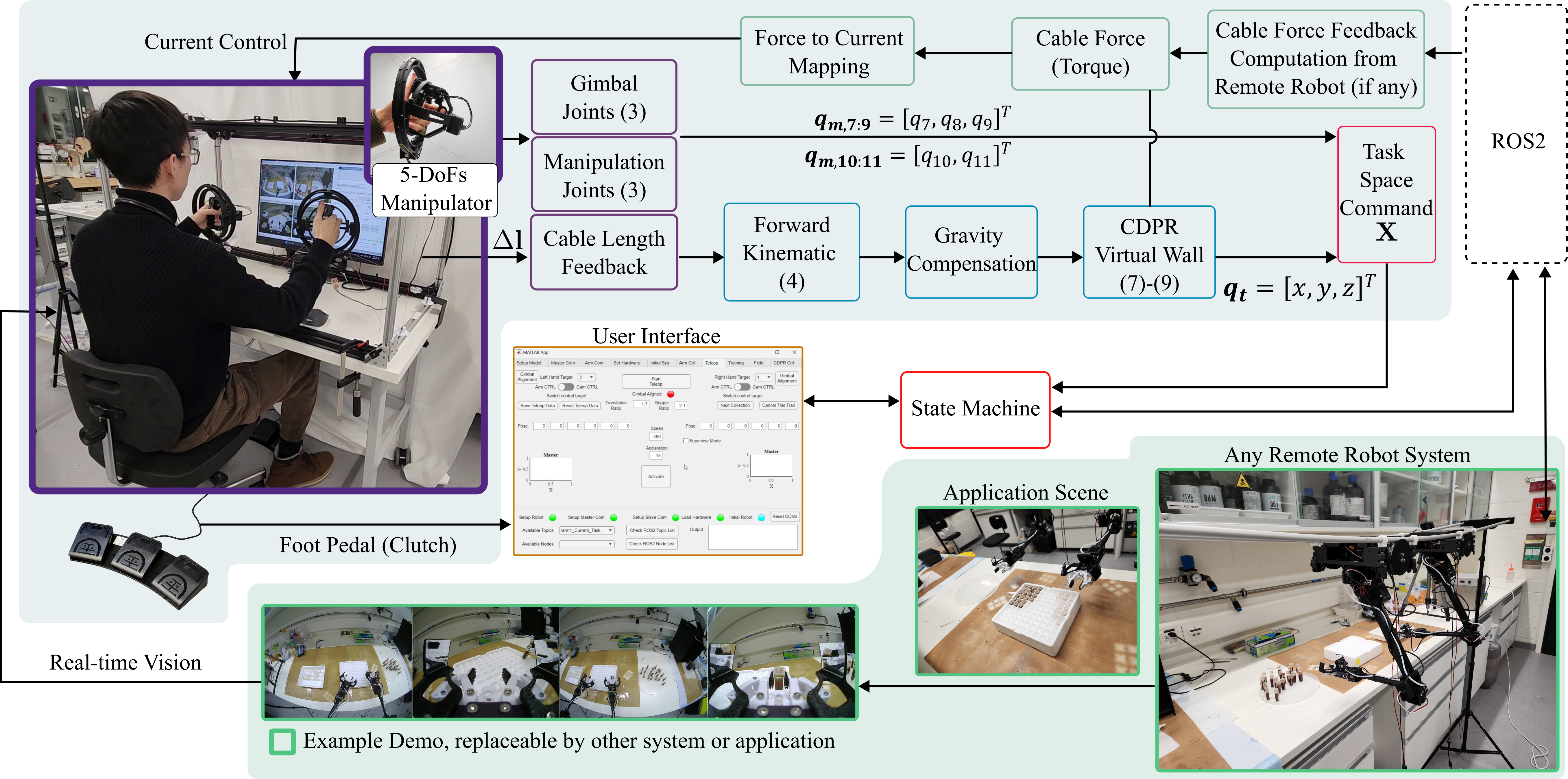}
    \caption{Control system schematic of the teleoperation system}
    \label{new_control_loop}
\end{figure*}

Additionally, any virtual fixture or feedback from the teleoperated robot can be transmitted to the user. Notably, haptic feedback for orientation, gripper, and rotational motions has not been implemented in this demonstration. However, the design and implementation of virtual fixtures and feedback are not the main focus of this work.

\subsection{Actuator Control Mode}
There are typically two modes using in the system: Position mode and Current (Force) mode. After the system is initialized to its reference position $\mathbf{q}_{t,0}$, current mode is majorly used to keep the cable in tension. Then the FK in Section II.C can be applied whist providing any force feedback to the user. For the cables, the position mode is rarely used. 

\section{Hardware Design and Application Setup}

In the meantime, the gimbal manipulator typically operates in two modes: position mode and torque mode (Or current mode in the proposed system). The position mode is only used to align the target follower's orientation. In torque mode, the manipulator usually runs in low-torque mode to enable gravity compensation, backdrivability and provide instance orientation haptic force. This operation mode prevents the user from rotating the CDPR’s orientation, as all torques are absorbed by the gimbal actuators, allowing only translational forces to be transferred to the CDPR. Additionally, a higher torque mode is applied only for haptic feedback when reaching orientation or joint limits.

\subsection{Communication Setup}
The teleoperation system runs on ROS2 and serial communication. 
\Cref{new_control_loop} summarized the control system's schematic during the teleoperation experiment.
For the master controller, the motors are communicating using the U2D2 of Dynamixel. The master computer is powered by an 11th Gen i7-1165G7 @ 2.80GHz with 16GB RAM. The master computer calculated the kinematic and force control for both left and right hand CDPRs simultaneously using MATLAB R2024a with a Graphical user interface(GUI). 
In the experiments, the tailored teleoperated robotic arms operate on a Raspberry Pi 5 with 4GB RAM. 
Wireless communication between the master and teleoperated  systems has an delay ranging from 50-100 ms. The mass of the gimbal manipulator is 328g. The cable used in the system is Dingbear 1093Yd.
\begin{figure}[h]
    \centering
\includegraphics[width=0.8\columnwidth]{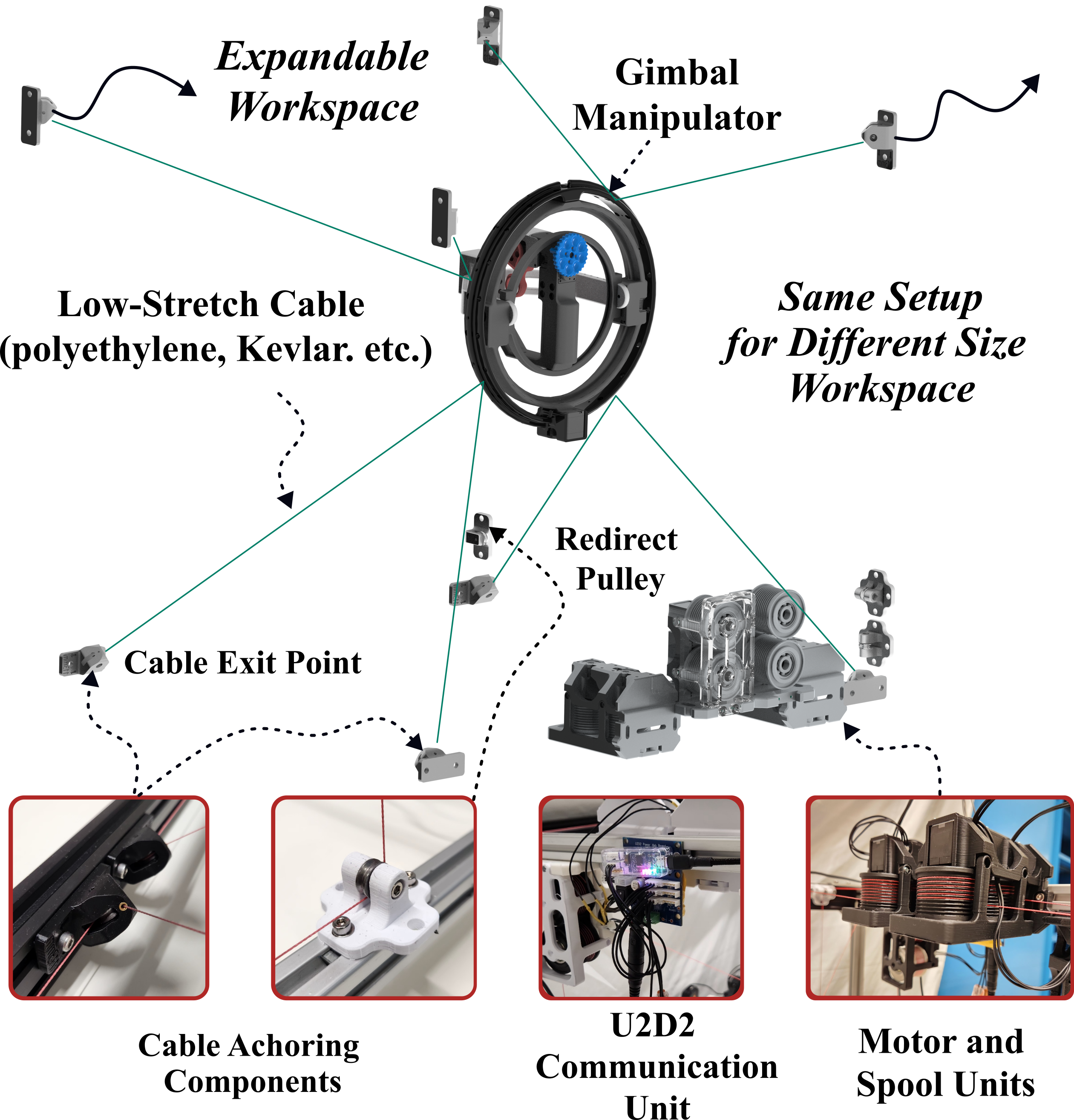}
    \caption{Hardware Design and Four Main Components of the system}
    \label{hardware_setup}
\end{figure}

\subsection{Actuators Selection and Hardware components}

Motors were selected for their speed and backdrivability. CDPRs require high RPM low back-drive force actuators for quick response and precise spatial feedback, while the gimbal manipulator prioritizes ease of movement. 
The system uses eight XL330-M077-T Dynamixel motors (0.228 Nm, 456 rpm) on each side of the CDPRs. Their backdrivability enables smooth end-effector movement with low-current control during teleoperation, providing adjustable force feedback to the user. The gimbal manipulator uses five XC330-M181-T Dynamixel motors (0.66 Nm, 155 rpm), switching between position and current modes depending on the robot's state.

Additional components include 3D-printed pulleys, motor mounts, spools, and an aluminum frame as shown in \Cref{hardware_setup}. Cable attachment positions were calibrated using the OptiTrack system, with a approximately $\pm$2 mm margin of error limited by the design.

\section{Experimental Results}
To demonstrate the application of the proposed master controller, a dual-arm robot is deployed to perform dexterous manipulation tasks using the 8-DoF robotic arm described in \cite{cheng2025cafes}. The robot's configuration, including the cable attachment points ${}^0\mathbf{A}_i$, is measured using the OptiTrack system with minimal measurement error. Before presenting the application scenarios, the performance of the robot is first evaluated. Over 9000 data points of position and orientation over the workspace are collected using the OptiTrack system for this analysis (with 1.2mm of errors as stated in the optitrack system).
\subsection{State-of-art Spatial Teleoperation Device}
\begin{table*}[ht]
    \centering
    \caption{Comparison of Commercial Haptic Devices and the proposed system. }
    \label{tab:haptic_devices}
    \resizebox{\textwidth}{!}{%
    \begin{tabular}{lccccccc}
        \toprule
        \textbf{Device} & \textbf{DOF} & \textbf{Trans. WS} ($10^{-3}m^3$) & \textbf{Rot. WS} ($^\circ$) & \textbf{Max Force} (N) & \textbf{Max Torque} (Nm) & \textbf{Weight} (kg) & \textbf{Add-on Input} \\
        \midrule
        Phantom Premium & 6 & 19.43 & $297^\circ \times 260^\circ \times 335^\circ$ & 8.5/37.5 & 0.515 & Unknown & - \\
        Geomagic Touch & 6 & 24.75 & Not Specified & 3.3 & 2.31 & 1.42 & - \\
        Geomagic Touch X & 6 & 14.57 & Not Specified & 7.9 & 2.35 & 3.26 & - \\
        Omega.3 & 3 & 2.2 & - & 12 & - & Unknown & - \\
        Omega.6 & 6 & 2.2 & $240^\circ \times 142^\circ \times 320^\circ$ & 12 & Not Actuated & Unknown & - \\
        Omega.7 & 7 & 2.2 & $240^\circ \times 140^\circ \times 320^\circ$ & 12 & Not Actuated & Unknown & Gripper (Pinching) \\
        Sigma.7 & 7 & 3.69 & $235^\circ \times 140^\circ \times 200^\circ$ & 20 & 0.4 & Unknown & Gripper (Pinching) \\
        Lambda.7 & 7 & 7.69 & $180^\circ \times 140^\circ \times 290^\circ$ & 20 & 0.4 & Unknown & - \\
        Delta.3 & 3 & 21.24 & - & 20 & - & Unknown & - \\
        Novint Falcon & 3 & 1.1 & - & Unknown & Unknown & 2.8 & Buttons (Grip) \\
        HapticMaster & 3 & 80 & - & 250 & - & Unknown & - \\
        Desktop 3D/6D & 3(6) & 45.76 & $100^\circ \times 200^\circ \times 300^\circ$ & 10 & N.A./0.8 & Unknown & - \\
        Virtuose 3D/6D & 3(6) & 781.85 & $120^\circ \times 330^\circ \times 270^\circ$ & 35 & 3.1/5 & 12 & Sensor Handle \\
        HD2 (Quanser) & 6 & 70 & $180^\circ \times 180^\circ$ Continuous & 19.71 & 1.72 & 22 & - \\
        Maglev 200 & 6 (7) & 0.00724 & $\pm 8^\circ$ & 40 & 3.6 & 18 & Handle (2 Btn) \\
        \midrule
\textbf{Proposed System} & \textbf{8} & \textbf{30 (35)$^1$} & $\bm{360^\circ \times 170^\circ \times 170^\circ}$ & \textbf{25} & \textbf{1.1} & \textbf{0.38 (10)$^2$} & \textbf{Gripper + Pedal} \\        
        \bottomrule
    \end{tabular}%
    } 
\footnotesize
\parbox[t]{\linewidth}{%
    \raggedright
    $^1$ 30 and 35 denotes the zero- and non-zero-orientation workspace, respectively, expandable by adjusting anchoring points without changing actuators.\\
    $^2$ 0.38 and 10 denote the weight of one end-effector and the entire setup, respectively.
}
\end{table*}
\Cref{tab:haptic_devices} summarizes existing commercial haptic devices and compares them with the proposed robot. The proposed system offers a larger workspace and more degrees of freedom (DoFs) than most commercial devices. Notably, the proposed system is easily expandable to a larger-scale workspace using the same hardware components, with only the anchoring positions needing adjustment. The current system is designed and optimized for the comfortable reach area of the human arm.

\subsection{Workspace and Virtual Fixture of the Master}
\begin{figure}[h]
    \centering    \includegraphics[width=0.9\columnwidth]{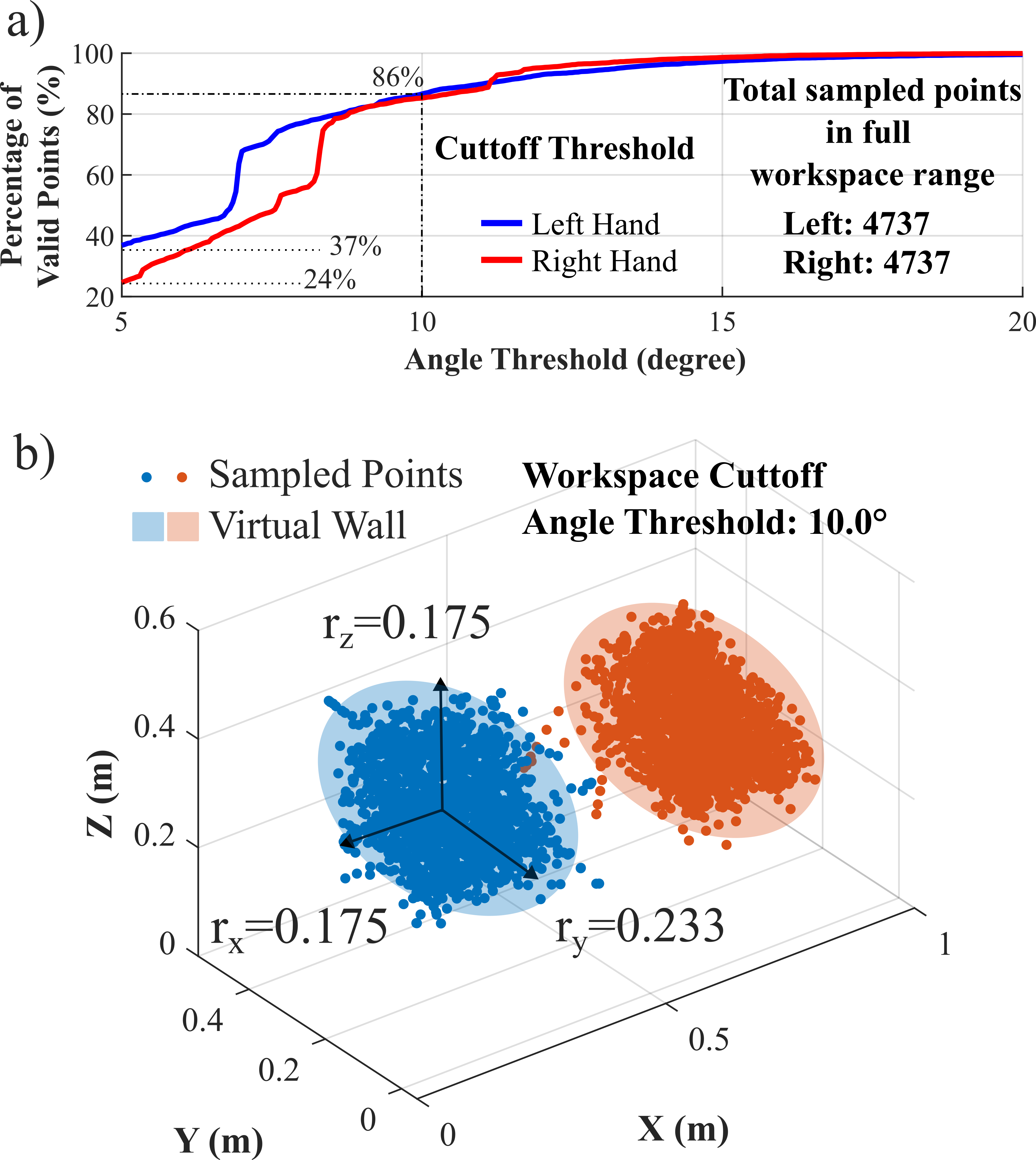}
    \caption{
    a) Sampled workspace points about the passive rotation (in x, y, z axis) generated across entire workspace. 98\% workspace are within the 15$\circ$ range. 
    b) Workspace (unit: m) of both CDPR master systems. The virtual wall is formed by an ellipsoid ($a=b=0.175 m, c=0.233 m$). However, the software clutch allows infinite extension by re-initializing the control reference.}
    \label{vitrual_fxiture}
\end{figure}

The used and overall workspace volume are approximately $30 \times 10^{-3}~\mathrm{m}^3$ and $35 \times 10^{-3}~\mathrm{m}^3$ respectively, based on the current configuration of the cable attachment points. This workspace can be expanded if needed. However, since the system does not actively regulate cable tension due to the absence of additional tension sensors (future work), it cannot dynamically balance the end-effector orientation. As a result, orientation errors naturally occur across the workspace.

As shown in \Cref{vitrual_fxiture}a), within the measured workspace ($35 \times 10^{-3}~\mathrm{m}^3$), most poses maintain low orientation errors along the $x$, $y$, or $z$ axes, with 98.5\% of the poses exhibiting orientation deviations smaller than $15^\circ$. Although the orientation of the end-effector is not used for teleoperation in this study, a cutoff threshold of $10^\circ$ is selected to minimize potential position errors. A virtual wall is defined based on this threshold, as illustrated in \Cref{vitrual_fxiture}b).

As described in the previous section, the virtual wall is implemented as an ellipsoid due to the characteristics of the gimbal manipulator design. When the robot approaches or exceeds the boundary of this virtual fixture, vibration feedback is triggered, as described in \Cref{sec:hapticfeedback}. At the same time, the teleoperated robot will halt and remain stationary until the end-effector returns within the virtual workspace

\subsection{Position and Orientation Performance of the Controller}
\begin{figure}[h]
    \centering    \includegraphics[width=1\columnwidth]{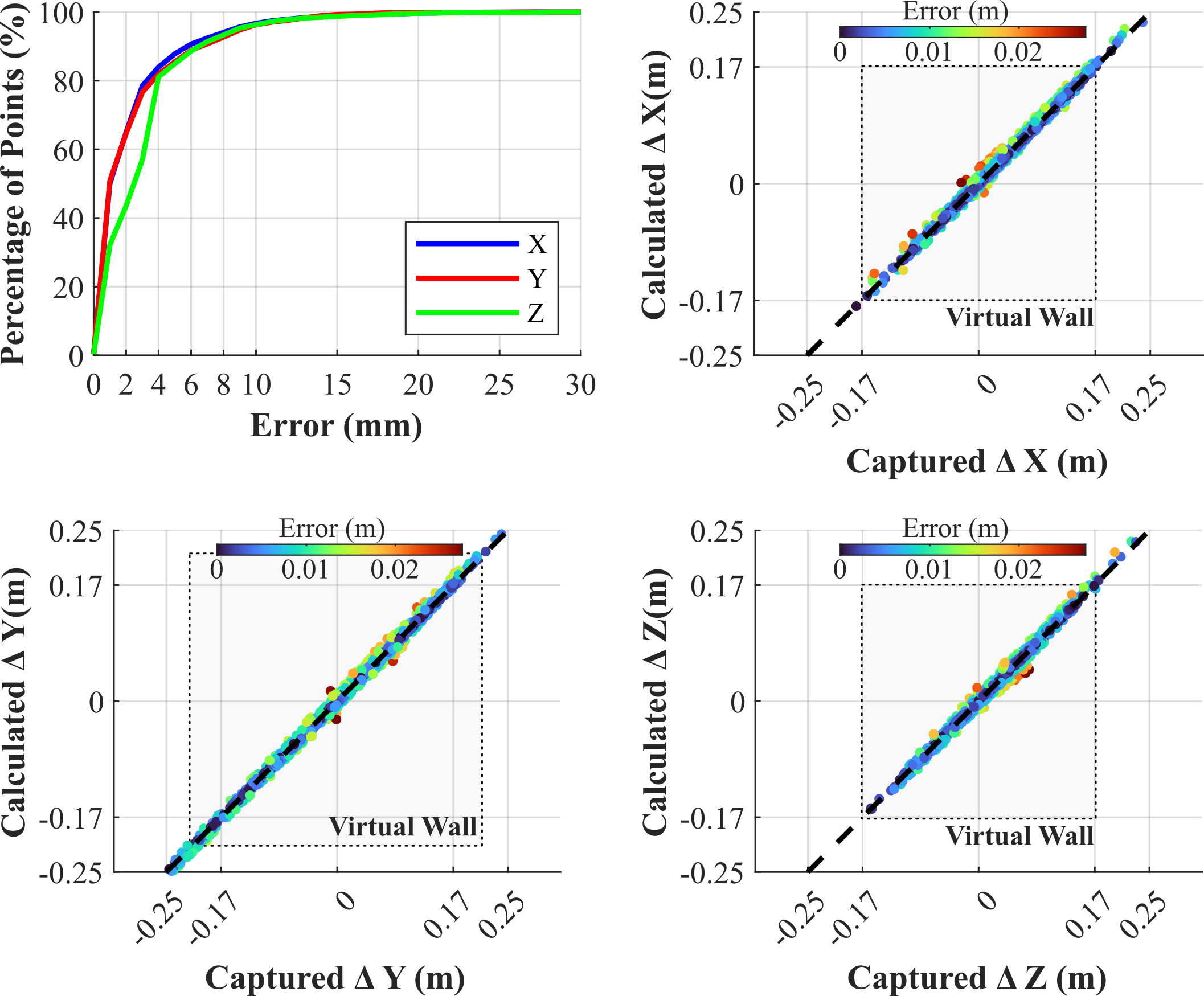}
    \caption{Relative position error  between the captured and calculated data with respect to the initial position of the robot}
    \label{displacement_comparison_m}
\end{figure}
To explore the decoupled behavior enabled by the proposed mechanical design, it is important to note that the translational degrees of freedom are handled by the CDPR, while the rotational degrees of freedom are controlled by the gimbal manipulator during the teleoperation. In other words, only the translational variables of the CDPR are used as the control input for the teleoperated robot. The system operates based on relative displacement, as described in \Cref{slave_pose}. The accuracy of the translational displacement is evaluated and presented in \Cref{displacement_comparison_m}. The results demonstrate that the system achieves high positional accuracy, considering the accumulated mechanical tolerances and measurement errors. More than 80\% of the displacement errors are within 4 mm.

Although the orientation of the CDPR is not utilized for teleoperation, examining its behavior still provides insight into the benefits of the decoupled design.

\begin{figure}[h]
    \centering
    \includegraphics[width=0.8\columnwidth]{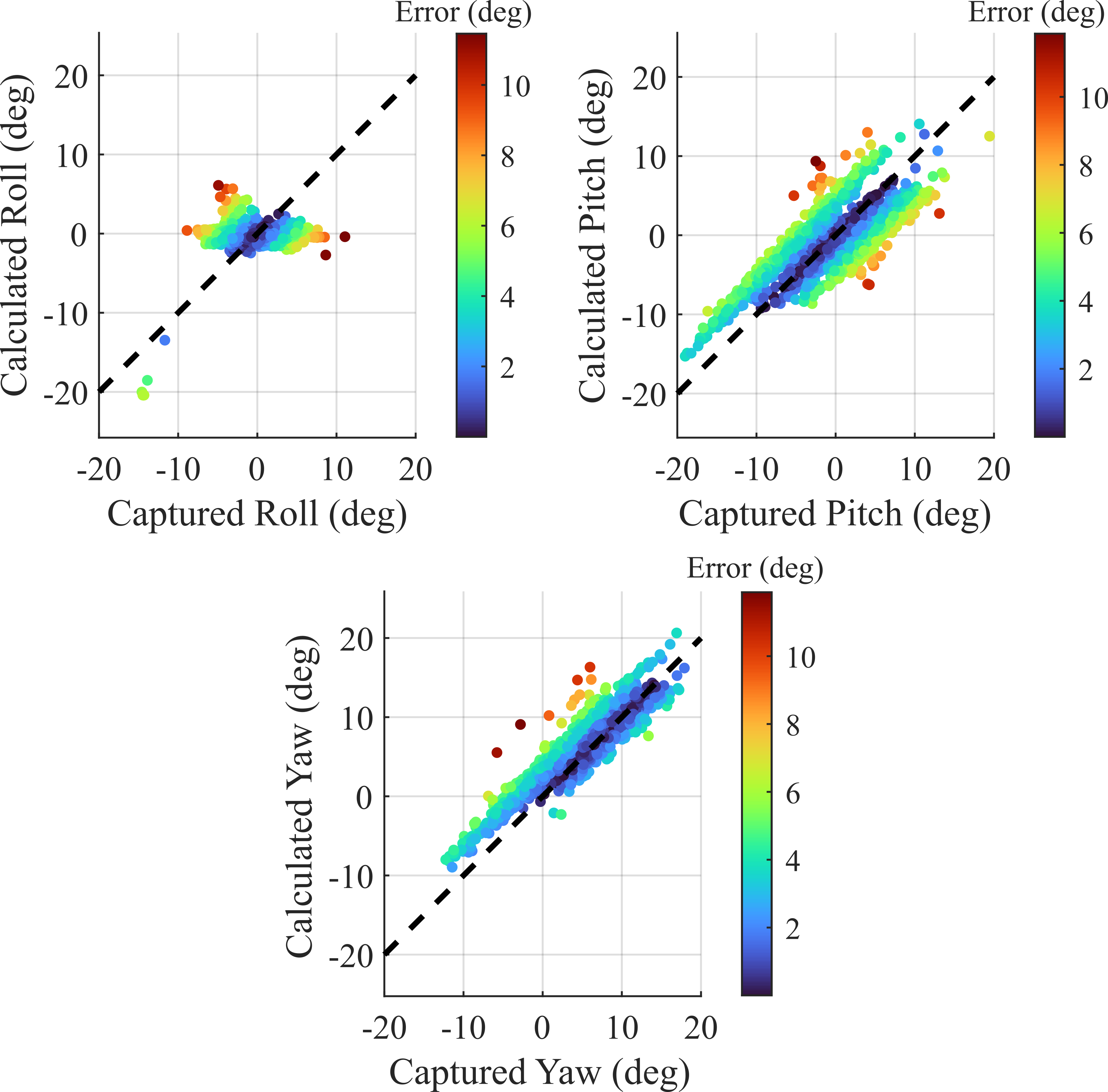}
    \caption{End-effector's (not the gimbal manipulator) orientation error between the captured and calculated data. Showing that the orientation of the end-effector is difficult to estimate with a random initial guess.}
    \label{orientation_comparison_m}
\end{figure}
When the operator change the angles of the gimbal manipulator, the end-effector naturally exhibits small orientation changes. This is due to a portion of the applied force being transmitted through the actuator’s static torque. As shown in \Cref{orientation_comparison_m}, the roll angle shows minimal variation compared to the pitch and yaw angles. This behavior is attributed to the cable attachment configuration $^1\mathbf{B}_i$, where the isotropic layout of the cables minimizes torque transmission along the roll axis. As a result, when the operator moves the end-effector, it is difficult to induce torque in the roll direction—especially if the gimbal manipulator is unlocked and free to move. In contrast, the pitch and yaw angles may exhibit arbitrary errors, typically within $10^\circ$ (that is why and where we set the threshold), due to the initial pose guess and computational limitations of the solver.

\begin{figure}[h]
    \centering
    \includegraphics[width=0.72\columnwidth]{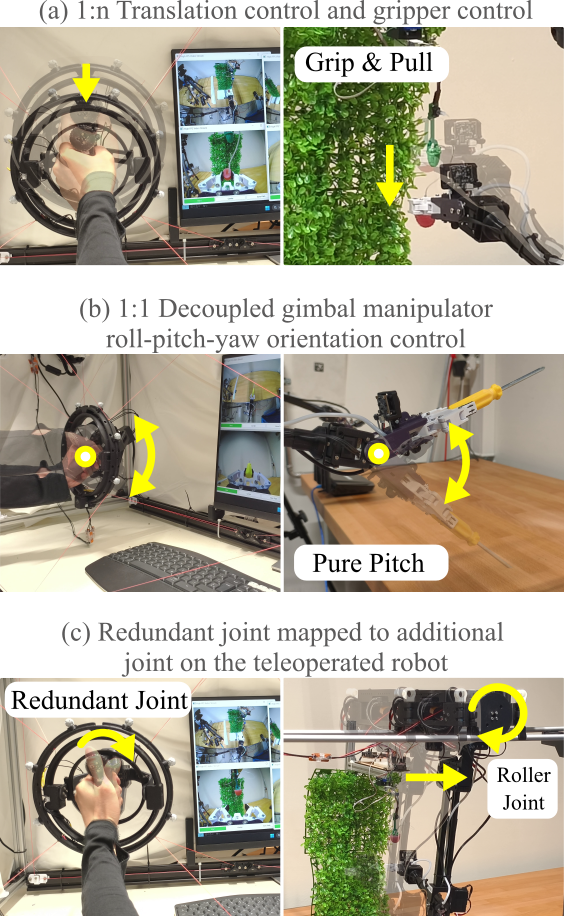}
    \caption{Motions from the master controller}
    \label{final_fig}
\end{figure}
Finally, \Cref{final_fig} illustrates the controlled degrees of freedom transferred from the master to an 8-DoF mobile robot \cite{cheng2025cafes}. The motion mappings include: $x$–$y$–$z$ translation with a $1:n$ scaling ratio, roll–pitch–yaw orientation with a $1:1$ mapping, gripper control, and redundant joint control. It is important to notices that the gimbal manipulator directly forward the joint angles to the teleoperated robot.

Additionally, it is important to clarify that the initial pose estimate is only coarsely defined, with the orientation assumed to be zero in the first iteration. In real applications, refining the initial guess can significantly reduce estimation errors and improve overall accuracy.

\subsection{Haptic Feedback Demo}
\begin{figure}[h]
    \centering
    \includegraphics[width=1\columnwidth]{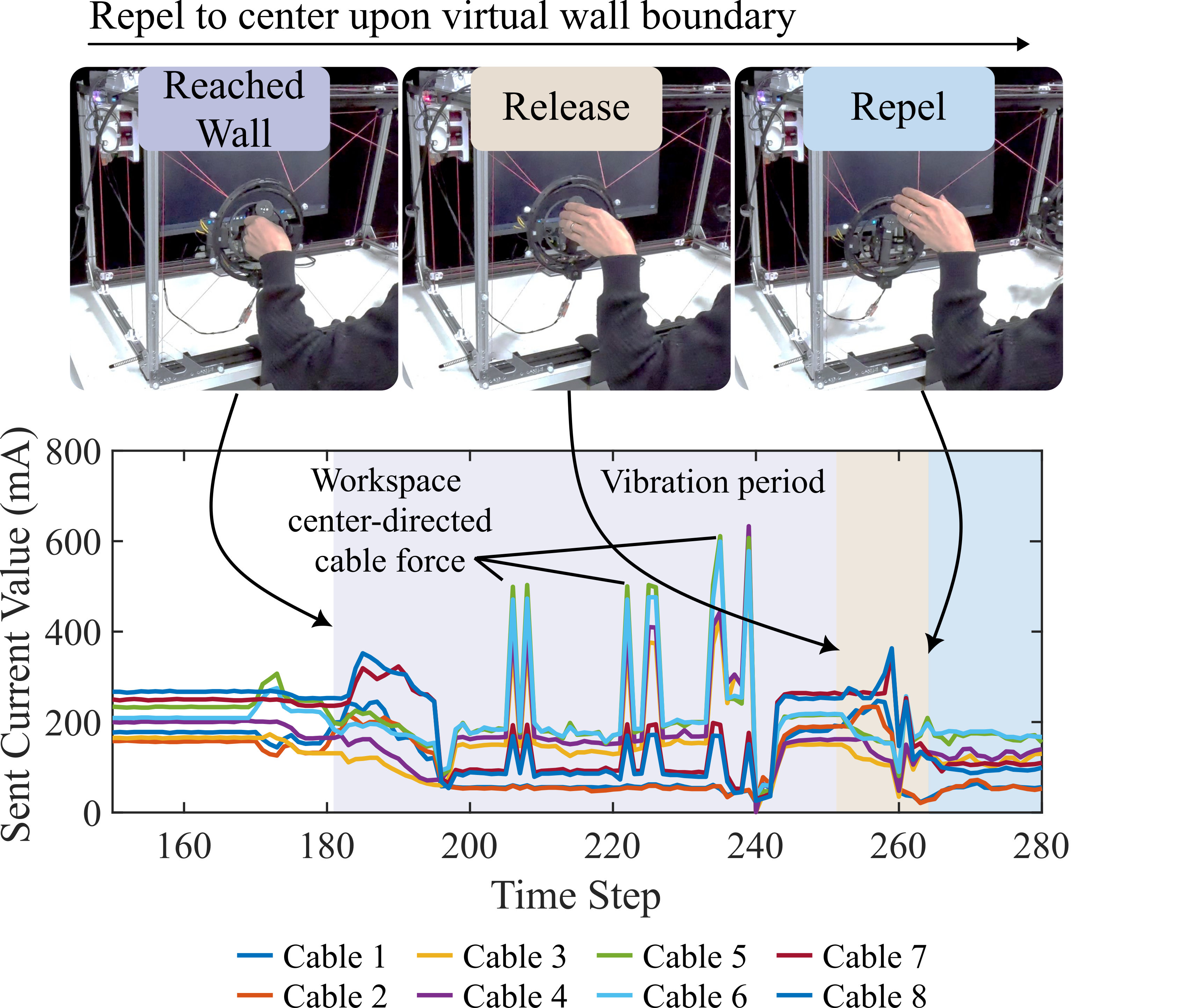}
    \caption{Sampled Haptic Feedback Demo: When the manipulator approaches the virtual wall, a vibration-based haptic feedback is triggered, generating a continuous task-space force directed toward the center of the ellipsoid. This force persists until the end-effector returns inside the virtual boundary.}    \label{virtual_wall_bouncing}
\end{figure}

To demonstrate the haptic feedback behavior of the robot, \Cref{virtual_wall_bouncing} presents an example where the feedback is transmitted to the operator within approximately 2 seconds. When the operator reaches the virtual wall, the cable forces are calculated to generate a task-space force directed toward the center of the workspace. The three periodic pulses shown in \Cref{virtual_wall_bouncing} represent the current commands sent to the robot.

It is also worth noting that the cable force becomes lower after the robot is repelled toward the center. This occurs because maintaining the robot closer to the center requires less effort than when it is near the workspace boundary. Although not explicitly described in the methodology, a simple gravity compensation have been deployed by solving the robot’s equations of motion.

\subsection{Manipulation tasks with the teleoperated robot}

\begin{figure}[h]
    \centering
    \includegraphics[width=0.9\columnwidth]{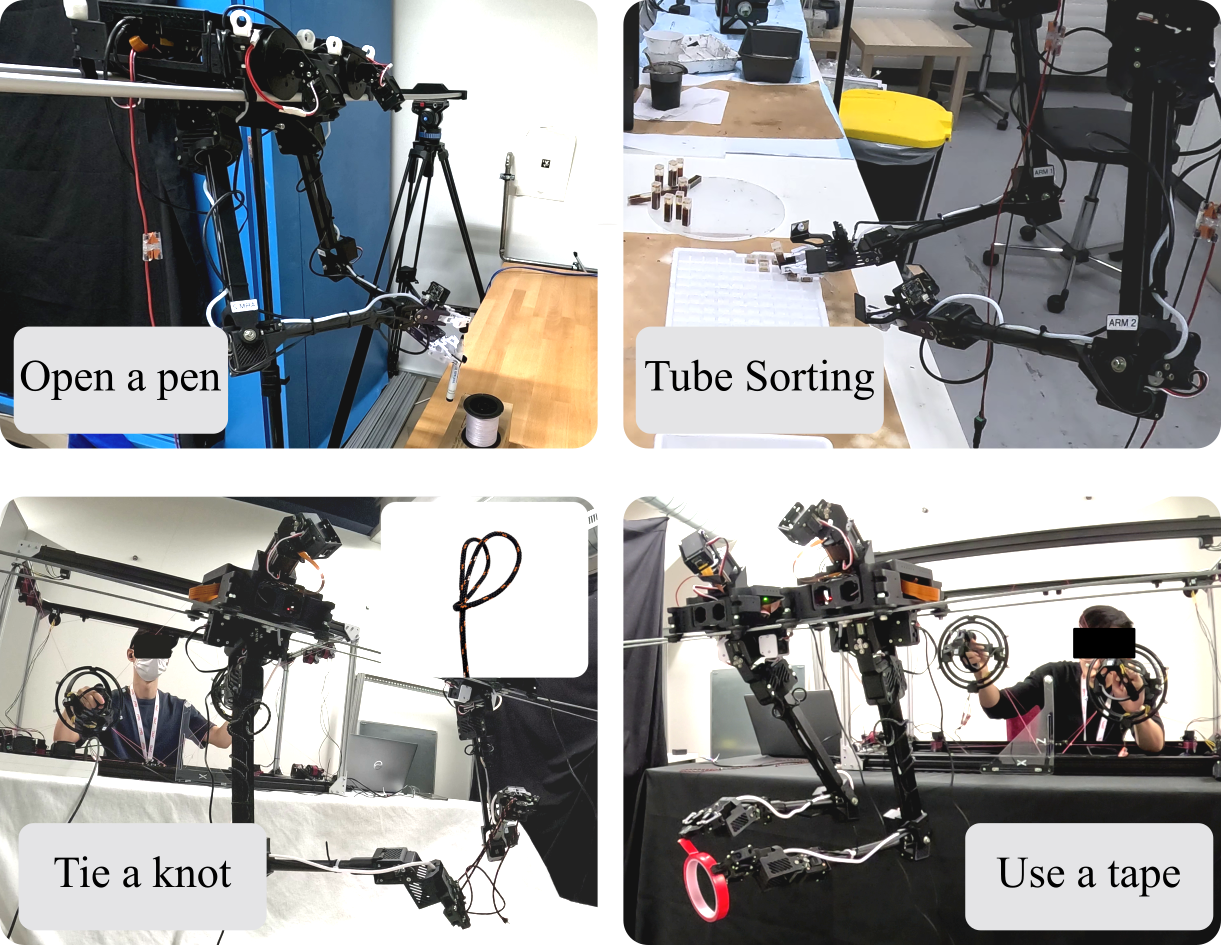}
    \caption{Bimanual tasks using the proposed system as teleoperation master controller}
    \label{example_cases}
\end{figure}

Finally, to demonstrate that the proposed system is capable of handling complex tasks such as pick-and-place and dexterous manipulation, \Cref{example_cases} presents several application examples using the master controller. These include opening a pen, sorting tubes, tying a knot, and using adhesive tape. Such demonstrations highlight the system’s versatility and precise control capabilities. Furthermore, the master controller is compatible with other high-DoF robots, including the UR5, Franka Emika, and even humanoid robots.
\section{{Conclusion and Future Works}}\label{sec:conclusion}
This paper proposes a design for an 8-DoF cable-driven parallel robot (CDPR) to be used as a teleoperation master controller. The goal of this work is to develop a low-cost yet highly accurate master interface. The design decouples the translational and rotational degrees of freedom (DoFs), addressing the issue of limited orientation workspace while leveraging the inherent advantage of a large translational workspace provided by CDPRs. Compared to most existing systems, the proposed master controller features an expandable workspace that includes both translation and orientation DoFs.

Experimental evaluation using an OptiTrack system demonstrates high control accuracy in the translational DoFs. For orientation, the system utilizes the output from a gimbal mechanism rather than relying solely on the CDPR. The accuracy of the system can be further improved by integrating force sensors, refining mechanical components to reduce tolerance errors, and incorporating an IMU to enhance orientation measurement and reduce CDPR-related orientation errors. Additionally, the actuators can be further enhanced with force-controlled mechanisms to provide more accurate haptic feedback.
\ifCLASSOPTIONcaptionsoff
  \newpage
\fi
\bibliographystyle{IEEEtran}
\bibliography{nb}

\end{document}